%% file: main.tex
\documentclass[10pt,twocolumn,letterpaper]{article}

\usepackage{iccv}              


\definecolor{iccvblue}{rgb}{0.21,0.49,0.74}
\usepackage[pagebackref,breaklinks,colorlinks,allcolors=iccvblue]{hyperref}

\usepackage{multirow}
\usepackage{microtype}
\usepackage{graphicx}
\usepackage{booktabs} 
\usepackage{colortbl}
\usepackage{amsmath}
\usepackage{bm}

\title{MGVQ: Could VQ-VAE Beat VAE? A Generalizable Tokenizer \\ with Multi-group Quantization}
\author{
  Mingkai Jia$^{1,2}$ \qquad
  Wei Yin$^{2}$\thanks{Corresponding author.} \qquad
  Xiaotao Hu$^{1,2}$ \qquad
  Jiaxin Guo$^{3}$ \qquad
  Xiaoyang Guo$^{2}$ \\
  Qian Zhang$^{2}$ \qquad
  Xiao-Xiao Long$^{4}$ \qquad
  Ping Tan$^{1}$\and
  $^1$The Hong Kong University of Science and Technology \quad $^2$Horizon Robotics \quad \\ $^3$The Chinese University of Hong Kong \quad $^4$Nanjing University
  }

\newcommand{\NickName}{\textit{MGVQ}}
\begin{document}

\makeatletter
\let\@oldmaketitle\@maketitle%
\renewcommand{\@maketitle}{\@oldmaketitle%
 \centering
    \includegraphics[width=0.9\linewidth]{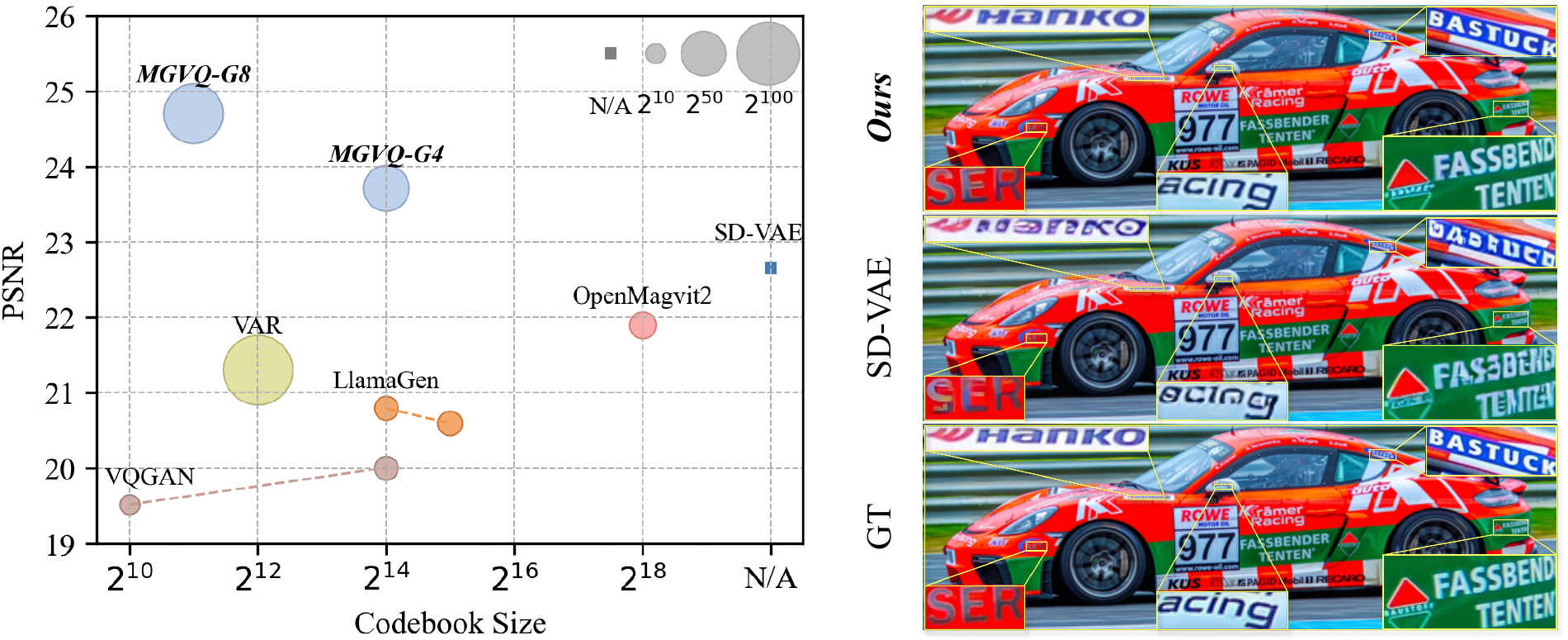}
     \captionof{figure}{
     \small \textbf{Reconstruction performance comparison between VQ-VAEs and SD-VAE on ImageNet 256×256 benchmark with $16\times $ downsampling.} The size of bubbles indicates the capacity, that is, the possibility of a token by sampling from the codebook. \textbf{\NickName-G8} with $8$ groups achieves a PSNR of $24.70$, evidently surpassing all others, with a large capacity of $2^{88}$. Qualitative results are illustrated where details are zoomed in for a better view.}
    \label{Fig: first page fig.}
    \bigskip}                   %
\makeatother
\maketitle

\input{sec/0_abstract}

\input{sec/1_intro}
\input{sec/2_relatedworks}

\input{sec/3_method}

\input{sec/4_exp.tex}

\input{sec/5_conclusion}

{
    \small
\bibliographystyle{ieeenat_fullname}
    \bibliography{main}
}

\end{document}

%% file: sec/0_abstract.tex
\begin{abstract}
Vector Quantized Variational Autoencoders (VQ-VAEs) are fundamental models that compress continuous visual data into discrete tokens. 
Existing methods have tried to improve the quantization strategy for better reconstruction quality, however, there still exists a large gap between VQ-VAEs and VAEs. 
To narrow this gap, we propose \NickName, a novel method to augment the representation capability of discrete codebooks,
facilitating easier optimization for codebooks and minimizing information loss, thereby enhancing reconstruction quality. Specifically, we propose to retain the latent dimension to preserve encoded features and incorporate a set of sub-codebooks for quantization. 
Furthermore, we construct comprehensive zero-shot benchmarks featuring resolutions of 512p and 2k to evaluate the reconstruction performance of existing methods rigorously.
\NickName~achieves the \textbf{state-of-the-art performance on both ImageNet and $8$ zero-shot benchmarks} across all VQ-VAEs.
Notably, compared with SD-VAE, we outperform them on ImageNet significantly, with \textbf{rFID $\textbf{0.49}$ v.s. $\textbf{0.91}$}, and achieve superior PSNR on all zero-shot benchmarks.
These results highlight the superiority of \NickName~in reconstruction and pave the way for preserving fidelity in HD image processing tasks. Code will be publicly available at \href{https://github.com/MKJia/MGVQ}{https://github.com/MKJia/MGVQ}.

\end{abstract}

%% file: sec/1_intro.tex
\section{Introduction}
\label{sec:intro}

\begin{figure*}[ht]
\begin{center}
\centerline{\includegraphics[width=2.1\columnwidth]{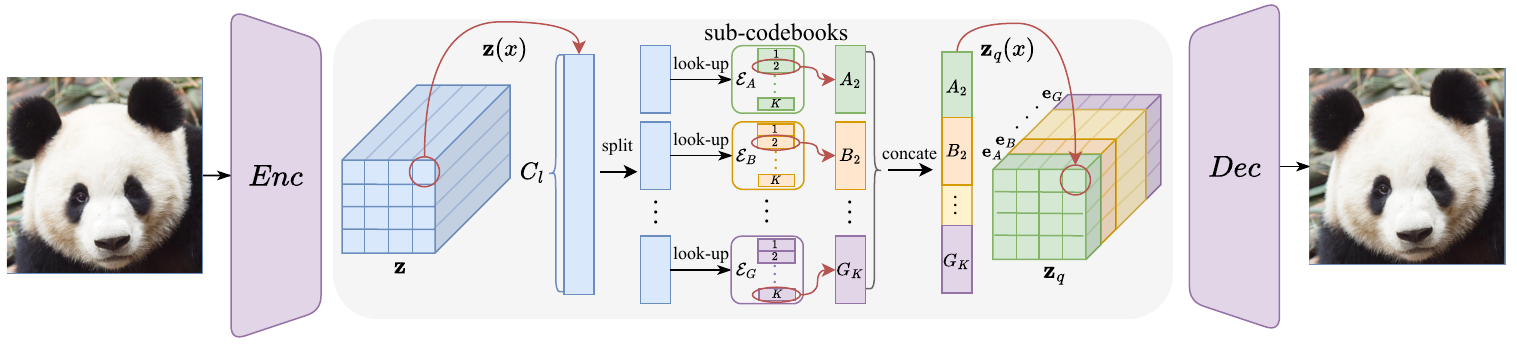}}
\caption{\textbf{An overview of \NickName~ framework.} \NickName~ keeps a larger dimension $C_l$ of latent $\mathbf{z}$ and split it into $G$ sub-tokens, where each sub-token is quantized individually with sub-codebook $\mathcal{E}_i$. Sub-tokens are then combined to compose $\mathbf{z}_q$ and for decoding. }
\label{fig:pipeline}
\end{center}
\vspace{-10mm}
\end{figure*}

Vector Quantization (VQ)~\cite{gray1984vectorquant} is a fundamental technique in computer vision that transforms continuous visual data into a set of  discrete set of tokens. By mapping high-dimensional data to a finite collection of learned codebook embeddings, VQ creates a compact and efficient latent space. This discrete representation not only reduces computational complexity but also retains critical semantic information, enabling more efficient data processing and analysis. As a key component in modern vision systems, VQ facilitates tasks such as image compression~\cite{agustsson2017soft,williams2020hierarchical,razavi2019generating,agarwal2025cosmos,yang2024cogvideox} and feature extraction~\cite{li2022unimo,liu2021cross,mao2021discrete,yu2021vector}, providing a robust foundation for a wide range of applications~\cite{zhang2023regularized,lee2022autoregressive,you2022locally,zheng2022movq}.

Despite recent advances, current tokenizers still suffer from significant information loss due to the limited representation capacity of codebooks. Consequently, Vector Quantized Variational Autoencoders (VQ-VAEs)~\cite{van2017vqvae, esser2021vqgan, yu2021vitvq, sun2024llamagen, luo2024oplfq, ma2025unitok, bai2024factorized} tend to exhibit inferior encoding-decoding performance compared to Variational Autoencoders (VAEs)~\cite{kingma2013vae}, often resulting in decoded images with substantial errors.
Simply increasing the number of feature dimensions and enlarging the codebook size typically leads to codebook collapse~\cite{takida2022sq}, where many code vectors remain inactive (receiving zero gradients during backpropagation). Thus, merely expanding codebook capacity fails to alleviate reconstruction errors.

Existing methods have explored various strategies to tackle these challenges. For instance, LlamaGen~\cite{sun2024llamagen} and Emu3~\cite{wang2024emu3} reduce the latent dimension to $8$ and $4$, respectively, to mitigate codebook collapse~\cite{takida2022sq} while expanding the codebook size to $16384$ and $32768$.
Other approaches, such as MAGVIT-v2~\cite{yu2023lfq}, introduce a lookup-free quantization (LFQ) strategy, reducing the code embedding dimension to zero and achieves an extra-large codebook size of $262144$.
While reducing the latent dimension enhanced the discrete latent space representation capability, it also results in information loss from the reduced dimension, ultimately degrading performance.
These trade-offs highlight the need for a more effective approach to expanding representation capacity while preserving information of encoded features and preventing codebook collapse.

To address these issues, we introduce \NickName, a novel method to enhance the representation capability of discrete codebooks. Unlike other methods that aggressively reduce dimensionality~\cite{yu2021vitvq,sun2024llamagen}, we preserve the latent dimension to maintain more expressive features while avoiding information loss. 
Specifically, we split the latent embeddings into sub-tokens, each mapped to an independent sub-codebook, facilitating easier optimization and larger representation capacity. Additionally, we employ a nested masking strategy during training to enforce the encoder to compress the image into the latent in an ordered manner.
By exponentially expanding the discrete latent space and reducing information loss, our \NickName~achieves higher fidelity and generalization ability across diverse benchmarks. Our experiments demonstrate that \NickName~outperforms existing VQ-VAE variants and significantly narrows the gap with VAE-based methods. Notably, \NickName~establishes new state-of-the-art results on ImageNet 256p and multiple zero-shot HD datasets, providing a promising direction for advancing the reconstruction performance of VQ-VAE models.

We summarize our contributions as follows.
\begin{enumerate}
    \item We propose \NickName, a novel method to enlarge the representation capability of discrete codebooks by over a billion-fold, aiming to narrow the reconstruction gap between VAE and VQ-VAE.
    \item We preserve the latent dimension and propose to use a group of sub-codebooks with a nested masking training strategy for quantization instead of a single unified one, achieving easier optimization for codebooks, larger representation capacity, and reduced information loss.
    \item We construct a zero-shot benchmark of 512p and 2k resolution to comprehensively evaluate the generalization ability for reconstruction. \NickName~realizes SOTA performance on both ImageNet 256p benchmark and $7$ zero-shot HD datasets. Note that \NickName~achieves rFID of $0.49$, significantly outperforming SD-VAE~\cite{rombach2022sd} of $0.91$.
\end{enumerate}

%% file: sec/2_relatedworks.tex
\section{Related Works}
\label{sec:related}

\subsection{Variational Autoencoder}
VAEs~\cite{kingma2013vae} have been instrumental in enhancing the performance of diffusion models, particularly in the context of image generation tasks. Recent advancements~\cite{zhao2025cvvae,zheng2024opensora,chen2024odvae,rombach2022sd,podell2023sdxl, gu2024dome, zhang2025epona} have significantly improved reconstruction quality and efficiency in these models.
One of the popular VAEs is the SD-VAE~\cite{rombach2022sd} model, fine-tuning the VAE decoder and trained on a combined dataset of billions of data for notable performance enhancements.
Building upon the foundations of SD-VAE~\cite{rombach2022sd}, SDXL-VAE~\cite{podell2023sdxl} introduces several architectural and training improvements. By employing larger batch sizes and incorporating exponential moving average weight tracking, it achieves superior performance in various reconstruction metrics compared to its predecessors.  

\subsection{Vector Quantized Variational Autoencoder}
VQ-VAEs extend traditional VAEs by discretizing latent representations into tokens from a learned codebook~\cite{van2017vqvae}. 
Based on VQ-VAE~\cite{van2017vqvae}, VQGAN~\cite{esser2021vqgan} further use a discriminator to keep perceptual quality at increased compression rate. 
Following the framework of VQ-VAE and VQGAN, several methods attempt to enhance the performance in many ways.
Some existing methods focus on enhancing the representation capability of discrete codes. 
ViT-VQGAN~\cite{yu2021vitvq} discovers that reducing dimension of latent space from $256$ to $32$ consistently improves reconstruction quality. Then LlamaGen~\cite{sun2024llamagen} further reduces dimension from $32$ to $8$, with an enlarged codebook of $16384$ that benefits the overall performance. 
However, they figure out that continuing to increase the codebook size (e.g. $32768$) may lead to a degradation of performance and lower usage of discrete codes. 
This shows the necessity to improve the usage of codes. 
To deal with this issue, SQ-VAE~\cite{takida2022sq} applies self-annealed stochastic quantization for greater codebook utilization. CVQ-VAE~\cite{zheng2023cvq} proposes a clustering VQ-VAE that selects encoded features as anchors to update unused codes. SimVQ~\cite{zhu2024addressing} reparameterizes the codes based on a learnable latent basis to optimize the entire linear space. 
However, a larger representation capacity remains a pressing need. 
Recent works turn to different types of quantization to a larger capacity of the discrete latent space. 
FSQ~\cite{mentzer2023fsq} replace VQ with a simple scheme termed finite scalar quantization, and MAGVIT-v2~\cite{yu2023lfq} proposes a similar lookup-free quantization (LFQ) strategy and achieves a super-large codebook size of $262144$ by reducing the dimension of code embedding to zero.
Nevertheless, they neglect the information loss with dimension downsampling, which is harmful for retaining the reconstruction quality.

%% file: sec/3_method.tex
\section{Method}

\label{sec:method}
\begin{figure}[t]
\begin{center}
\centerline{\includegraphics[width=1.0\columnwidth]{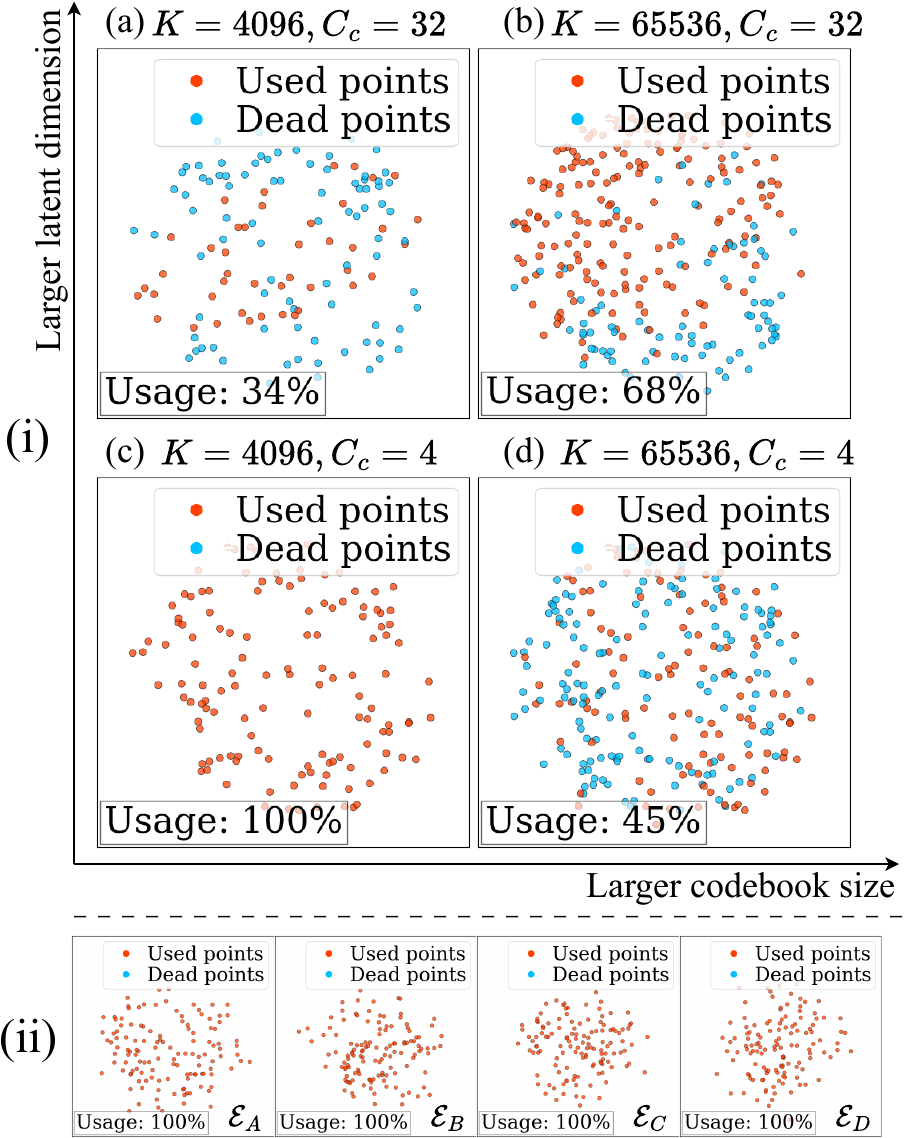}}
\caption{\textbf{(i) Codebook points of well-trained VQ-VAE models, that have different codebook sizes or latent dimensions.  (ii) Sub-codebook points in our proposed  \NickName~. The group size is 4.} \textcolor[rgb]{1.0, 0.295, 0.0}{Used points} are shown in red, while \textcolor[rgb]{0.0, 0.705, 1.0}{dead points} shown in blue. (i.a) A larger dimension and smaller size may lead to an anisotropic distribution and low usage. (i.b) A larger dimension and larger size show favor of certain directions, resulting in dead points in a specific area (lower right corner). (i.c) A smaller dimension and smaller size could be fully used without dead points. (i.d) A smaller dimension and larger size allow for a uniform spread, but there are more codes than needed, leading to some underutilization.}
\label{fig:deadpoints}
\end{center}
\vspace{-30pt}
\end{figure}

\subsection{Preliminary}
\noindent\textbf{VAE.} 
VAEs~\cite{kingma2013vae} are powerful deep generative models widely used to represent complex input data through a smaller latent space learned in an unsupervised manner~\cite{girin2020dvae}. 
A VAE consists of an encoder that parameterizes input data $\mathbf{x}\in \mathbb{R}^{H\times W\times 3}$ to a latent variable $\mathbf{z}\in \mathbb{R}^{\frac{H}{D} \times \frac{W}{D} \times C_l}$, where $H \times W$ are input spatial sizes, $D$ is the downsample scale, and $C_l$ is the latent dimension. A continuous latent $\mathbf{z}_s$ is sampled from the posterior distribution $q(\mathbf{z}|\mathbf{x})$ with a given prior distribution $q(\mathbf{z}) = \mathcal{N}(\mathbf{z}; \mathbf{0}, \mathbf{I})$, and VAE's decoder $p_\theta(\mathbf{x}|\mathbf{z})$ with parameters $\theta$ reconstructs the data~\cite{van2017vqvae}. 
The posterior distribution of a specific input data $\mathbf{x}$ can be approximated as a Gaussian distribution with approximately diagonal covariance matrix~\cite{kingma2013vae}:
\begin{equation}
    \log q_\phi(\mathbf{z}|\mathbf{x})=\log \mathcal{N}(\mathbf{z}|\mathbf{\mathbf{\mu}},\mathbf{\sigma}^2\mathbf{I}),
\end{equation}
where $\phi$ denotes variational parameters within the encoder, and $\mathbf{\mu}, \mathbf{\sigma}$ are the mean and standard deviation of the approximate posterior. 
The optimization objective is the evidence lower bound (ELBO), derived from Jensen’s inequality~\cite{kingma2019introduction}:
\begin{equation}
    \mathcal{L}_{\theta, \phi}(\mathbf{x})=\mathbb{E}_{q_\phi(\mathbf{z}|\mathbf{x})}[\log p_{\theta}(\mathbf{x},\mathbf{z})-\log q_{\phi}(\mathbf{z}|\mathbf{x})].
\end{equation}

\noindent\textbf{VQ-VAE.} 
Oord~\etal~\cite{van2017vqvae} reveals that training VAE easily faces posterior collapse and large variance, thus the Vector Quantized Variational Autoencoder (VQ-VAE) is proposed. Instead of sampling from the posterior distribution, it proposes to encode the inputs into discrete latents and remove the KL term. 
They set up a learnable codebook $\mathcal{E} =\{\mathbf{e}_i\}_{i=1}^K \subset \mathbb{R}^{K \times C_c}$, where $K$ is the number of learnable vectors, and $C_c$ is the vector's dimension~\cite{esser2021vqgan}. They employ a channel downsample to $C_c$ first and then use the nearest neighbor lookup method to map the encoder's latent $\mathbf{z}\in \mathbb{R}^{\frac{H}{n} \times \frac{W}{n} \times C_c}$ to the quantized latent $\mathbf{z}_q \in \mathbb{R}^{\frac{H}{n} \times \frac{W}{n} \times C_c}$:
\begin{equation}
\mathbf{z}_q(x) = \mathbf{e}_k, \quad \text{where} \quad k = \arg\min_{\mathbf{e}_i \in \mathcal{E}} \|\mathbf{z}(x) - \mathbf{e}_i\|_2.
\end{equation}
where $x$ represents a token.
The quantized latent $\mathbf{z}_q$ is fed to the decoder for reconstruction.
Note that in their implementations~\cite{sun2024llamagen, wang2024emu3}, they usually employ a small quantization dimension $C_c$ to avoid codebook collapse during training.

\subsection{What limits the reconstruction quality of VQ-VAE?}
\label{sec:key_factors}

\noindent\textbf{Information loss.}
Recent research on VAE~\cite{esser2024sd3, chen2024dcae, hansen2025vitok} reveals that a larger latent space dimension (i.e. $C_l$) significantly boosts reconstruction quality. For example, SD3~\cite{esser2024sd3} increases the latent space channel number from 4 (SD1~\cite{rombach2022sd}) to 16, while DC-AE~\cite{chen2024dcae} employs up to $128$ channels in their implementations. However, training codebook discretization easily face codebook collapse~\cite{takida2022sq, zheng2023cvq}, where codes do not receive gradients, resulting in unoptimized during training and unused in inference, also known as dead points. To avoid this, VQ-VAE employs a much lower latent dimension $C_c$. However, this may 
introduces noticeable information loss, leading to inferior reconstruction quality. 
For example, LlamaGen~\cite{sun2024llamagen} reduces the latent channel from $256$ to $8$ for quantization and Emu3~\cite{dai2023emu} use $4$ quantization channels.

\noindent\textbf{Limited latent representation capacity.} 
The representation capacity of a VQ-VAE denotes the sampling possibility from its codebooks. For example, LlamaGen's representation capacity can be denoted as 16384 with a codebook size of 16384. A larger representation capacity means a stronger reconstruction ability.
VQ-VAEs utilize a discrete latent space represented by a learned codebook $\mathcal{E}$ with limited vectors, while VAE encodes images into the continuous latent variables $\mathbf{z}$, resulting in a continuous latent space. Compared with VAE, such a finite codebook lags VQ-VAE's representation capacity, resulting in worse reconstruction results, especially for texture-complex images.  

This brings up the question: \emph{Can we simultaneously improve the codebook size $K$ and the quantized latent dimension $C_c$ to boost VQ-VAE's performance}?
Unfortunately, we observe that there is an inherent conflict between the latent dimension and enlarging the codebook size during training.
~\cref{fig:deadpoints} illustrates an experiment of code points. 
We trained 4 models with different codebook sizes or latent dimensions and showed their codebook points. Compared with the baseline (~\cref{fig:deadpoints} (c)),
~\cref{fig:deadpoints} (a)(b)(d) present that either enlarging the codebook size or increasing latent dimension size will cause significant dead codes, i.e. points not optimized in training and unused in inference. 
In contrast, the codebook usage of the baseline (~\cref{fig:deadpoints} (c)) is much higher.

\begin{table*}[t]
\centering
\setlength{\tabcolsep}{4.0pt} 
\begin{tabular}{lccccccccc}
\toprule
\multirow{2}{*}{\textbf{Method}} & 
\multirow{2}{*}{\textbf{Quantization}} &
\multirow{2}{*}{\textbf{Ratio}} &
\multirow{2}{*}{\textbf{Dim}} & 
\multirow{2}{*}{\textbf{Size}} & 
\multirow{2}{*}{\textbf{Capacity}} &
\multirow{2}{*}{\textbf{\shortstack{Codebook\\Usage}}} & 
\multicolumn{3}{c}{\textbf{ImageNet}} \\
\cmidrule(lr){8-10}& & & & & & & \textbf{rFID}$\downarrow$ & \textbf{PSNR}$\uparrow$ & \textbf{SSIM}$\uparrow$ \\
\midrule
 \rowcolor[gray]{.9}SD-VAE\textsuperscript{ukn.}~\cite{rombach2022sd} & - & 16& 16 & - & -&- & 0.91&22.65&0.697 \\
VQGAN~\cite{esser2021vqgan} & VQ & 16&256 & 1024 & $1024^1 =  2^{10}$ & -&8.30 & 19.51 & 0.614 \\
VQGAN~\cite{esser2021vqgan} & VQ & 16& 256 & 16384 & $16384^1 = 2^{14}$ & -& 4.99 & 20.00 & 0.629  \\
LlamaGen~\cite{sun2024llamagen} & VQ & 16& 8 & 16384 & $16384^1 = 2^{14}$ & 97\%& 2.19 & 20.79 & 0.675 \\
LlamaGen~\cite{sun2024llamagen} & VQ & 16& 8 & 32768 & $32768^1 = 2^{15}$ & 85\% & 2.26 & 20.59 & 0.663 \\
OpenMagvit2~\cite{luo2024oplfq} & LFQ & 16& 18 & 262144 & $262144^1 = 2^{18}$ & 100\%& 1.17 & 21.90 & - \\
VAR~\cite{tian2025visual} & Multi-scale VQ& 16 & 32 & 4096 & $4096^{10}=2^{120}$& - & - & 21.30 & 0.647  \\
\textbf{\NickName-G4} & VQ& 16 & 32 & 8192 $\times$ 4 & $8192^4=2^{52}$& 100\% & \underline{0.64} & \underline{23.71} & \underline{0.755}    \\
\textbf{\NickName-G8} & VQ & 16& 32 & 2048 $\times$ 8 & $2048^8=2^{88}$& 100\% & \textbf{0.49} & \textbf{24.70} & \textbf{0.787}    \\

\midrule
\rowcolor[gray]{.9}SD-VAE\textsuperscript{ukn.}~\cite{rombach2022sd} & - &8& 4 & - & -& - & 0.74 & 25.68 & 0.820 \\
\rowcolor[gray]{.9}SDXL-VAE\textsuperscript{ukn.}~\cite{podell2023sdxl} & -&8 & 4 & - & -& -& 0.68 & 26.04 & 0.834 \\
 VQGAN\textsuperscript{oim.}~\cite{esser2021vqgan} & VQ &8& 4 & 256 & $256^1=2^{8}$ & -& 1.44 & 22.63 & 0.737 \\
VQGAN\textsuperscript{oim.}~\cite{esser2021vqgan} & VQ &8& 4 & 16384 & $16384^1=2^{14}$& - & 1.19 & 23.38 & 0.762\\
LlamaGen~\cite{sun2024llamagen} & VQ &8& 8 & 16384 & $16384^1=2^{14}$& - & 0.59 & 24.45 & 0.813   \\
OpenMagvit2~\cite{luo2024oplfq} & LFQ &8& 18 & 262144 & $262144^1 = 2^{18}$& 100\% & {0.34} & 27.02 & -  \\
\textbf{\NickName-G4} & VQ &8& 32 & 8192 $\times$ 4 & $8192^4=2^{52}$& 100\% & \underline{0.31} & \underline{28.42} & \underline{0.898}    \\
\textbf{\NickName-G8} & VQ &8& 32 & 2048 $\times$ 8 &  $2048^8=2^{88}$& 100\%  & \textbf{0.27} & \textbf{29.96} & \textbf{0.918}   \\
\bottomrule
\end{tabular}
\caption{\textbf{Reconstruction evaluation on 256$\times$256 ImageNet benchmark.} \NickName-G4 indicates our method with $4$ sub-codebooks, and \NickName-G8 indicates $8$ sub-codebooks. All models are trained on ImageNet except ``oim.'' is on OpenImage, ``ukn.'' is unknown training data. \colorbox[gray]{.9}{Gray} denotes continuous tokenizers. }
\label{tab:recon_in256}
\end{table*}
\begin{figure}[t]
\begin{center}
\centerline{\includegraphics[width=1.0\columnwidth]{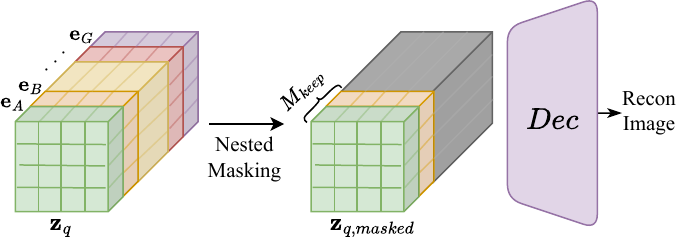}}
\caption{\textbf{The process of nested masking.} \textcolor[rgb]{0.44, 0.44, 0.44}{Gray blocks} represents the masked last tokens, and other colors show active sub-groups.}
\label{fig:nested}
\end{center}
\vspace{-35pt}
\end{figure}

\begin{figure*}[t]
\begin{center}
\centerline{\includegraphics[width=2.0\columnwidth]{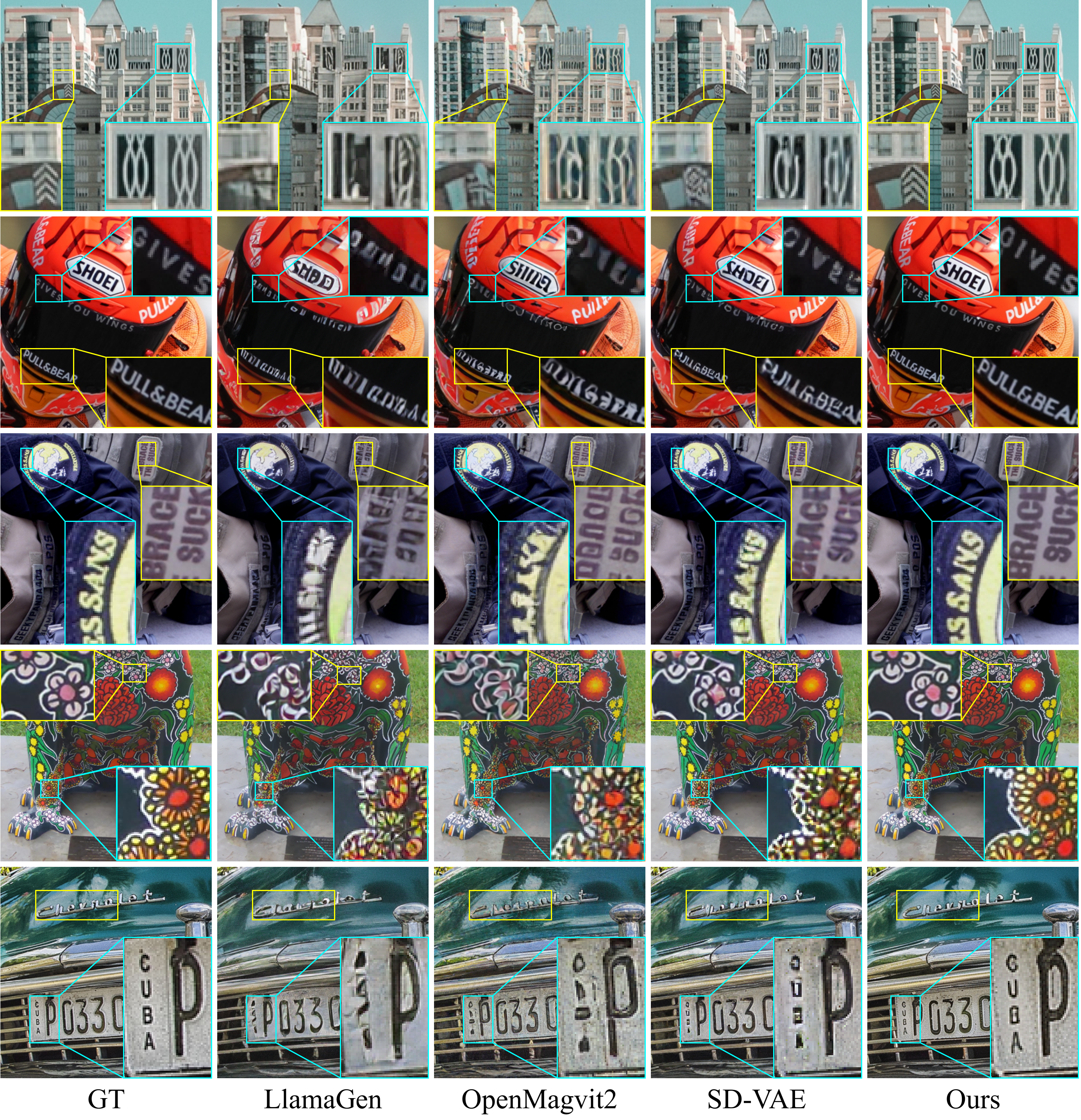}}
\caption{\textbf{Qualitative reconstruction images with $16\times$ downsampling on 2560 $\times$ 1440 UHDBench dataset.} We crop a $360\times 360$ sub-region, and zoom in \textit{detailed textures} using \textcolor{blue}{blue} and \textcolor{yellow}{yellow} for better view.}
\label{fig:qual}
\end{center}
\vspace{-30pt}
\end{figure*}

\begin{table*}[t]
\centering
\setlength{\tabcolsep}{1.0pt} 
\begin{tabular}{lcccccccccccccc}
\toprule
\multirow{2}{*}{\textbf{Method}} & \multicolumn{2}{c}{\textbf{CelebA}} & \multicolumn{2}{c}{\textbf{DAVIS}} & \multicolumn{2}{c}{\textbf{TextOCR}} & \multicolumn{2}{c}{\textbf{VFHQ}} & \multicolumn{2}{c}{\textbf{Spring}} & \multicolumn{2}{c}{\textbf{ENeRF}} & \multicolumn{2}{c}{\textbf{Ranking}} \\
\cmidrule(lr){2-3} \cmidrule(lr){4-5} \cmidrule(lr){6-7} \cmidrule(lr){8-9} \cmidrule(lr){10-11} \cmidrule(lr){12-13} \cmidrule(lr){14-15}&\textbf{rFID}$\downarrow$ & \textbf{PSNR}$\uparrow$ & \textbf{rFID}$\downarrow$ & \textbf{PSNR}$\uparrow$ & \textbf{rFID}$\downarrow$ & \textbf{PSNR}$\uparrow$ & \textbf{rFID}$\downarrow$ & \textbf{PSNR}$\uparrow$ & \textbf{rFID}$\downarrow$ & \textbf{PSNR}$\uparrow$ & \textbf{rFID}$\downarrow$ & \textbf{PSNR}$\uparrow$ & \textbf{rFID} & \textbf{PSNR} \\
\midrule
\rowcolor[gray]{.9}SD-VAE~\cite{rombach2022sd}    & \underline{0.73} & \underline{30.81} & \textbf{9.30} & \underline{23.37} & \underline{2.38} & \underline{26.30} & \underline{4.71} & \underline{30.82} & \textbf{19.29} & \underline{25.26} & \textbf{5.21} & \underline{25.13} & \textbf{1.5} & \underline{2.0} \\ 
\midrule

VQGAN~\cite{esser2021vqgan}        & 7.19 & 25.43 & 26.05 & 20.23 & 6.49 & 21.40 & 18.58 & 25.97 & 52.65 & 22.34 & 19.26 & 21.62 & 7.0 & 7.0 \\
 LlamaGen~\cite{sun2024llamagen} & 2.45 & 26.09 & 19.96 & 21.20 & 4.94 & 22.47 & 12.15 & 26.46 & 37.49 & 23.05 & 13.35 & 22.52 & 5.7 & 6.0 \\ 
   OpenMagvit2~\cite{luo2024oplfq} & {1.35} & {27.41} & {15.56} & {21.64} & {4.19} & {23.42} & {7.27} & {27.85} & {32.31} & {23.48} & 8.91 & {23.03} & {3.2} & {4.0} \\
    VAR~\cite{tian2025visual} & 1.73 & 26.50 & 17.84 & 21.41 & 4.49 & 23.19 & 8.46 & 27.22 & 34.25 & 23.26 & {8.89} & 22.98 & 4.0 & 5.0 \\ 
      \NickName-G4 & \textbf{0.69} & \textbf{31.25} & \underline{9.52} & \textbf{26.20} & \textbf{2.30} & \textbf{27.95} & \textbf{3.42} & \textbf{31.96} & \underline{22.30} & \textbf{26.76} & \underline{5.83} & \textbf{26.98} & \textbf{1.5} & \textbf{1.0} \\
\bottomrule
\end{tabular}
\caption{\textbf{Zero-shot reconstruction evaluation with a downsample ratio of 16 on 512$\times$512 datasets.} \colorbox[gray]{.9}{Gray} denotes continuous tokenizers. \NickName-G4 indicates our method with $4$ sub-codebooks. }
\label{tab:recon_zero512}
\vspace{-5mm}
\end{table*}

\begin{figure*}[t]
\begin{center}
\centerline{\includegraphics[width=2.0\columnwidth]{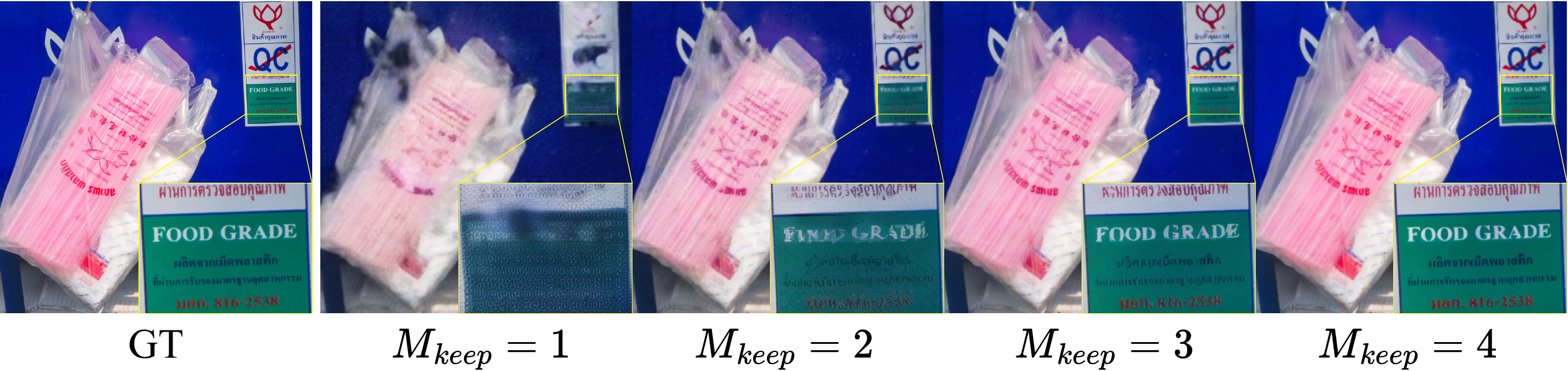}}
\caption{\textbf{Qualitative ablation of our nested masking strategy.} The first figure is the input image. Others show reconstruction results using first $M_{keep}$ sub-groups for $G=4$ from left to right. The output images gradually gain detailed textures using more sub-groups. The \textcolor{yellow}{yellow} box zooms in \textit{letters} for better view.}
\label{fig: qual_ab_c2f}
\end{center}
\vspace{-35pt}
\end{figure*}
\subsection{MGVQ}
\label{sec:multi-group}

To tackle the aforementioned conflicts and narrow the reconstruction gap between VAE and VQ-VAE, we propose \NickName, a novel method to enlarge the representation capability of discrete codebooks. To enhance the VQ-VAE performance, our insight is that the key point is increasing the representation capacity instead of the codebook size. Therefore, we attempt to enlarge the representation capacity but maintain a small codebook size for easy optimization.
~\cref{fig:pipeline} presents our pipeline. Following VQGAN~\cite{esser2021vqgan}, we employ a convolution-based backbone to extract features $\mathbf{z}\in \mathbb{R}^{\frac{H}{D} \times \frac{W}{D} \times C_l}$. 
In contrast to existing methods reducing the latent channel to a small one before quantization, we use a larger latent dimension $C_l$ than them to retain as much encoded feature as possible. This can avoid information loss. 
Instead of using a single unified codebook, we propose to use a group of sub-codebooks for quantization. Specially, for each latent embedding $\mathbf{z}(x) \in \mathbb{R}^{C_l}$, 
we uniformly split it to $G$ sub-tokens $\{\mathbf{z}(x_{i})\}_{i=1}^{G}$, where each sub-token has the same dimension $\frac{C_l}{G}$. 
Correspondingly, we set up $G$ non-shared sub-codebooks, each with the same size $K$.
Each sub-token is quantized individually using its associated sub-codebook, 
$\mathcal{E}_i =\{\mathbf{e}_{i}^{j}\}_{j=1}^K \subset \mathbb{R}^{K \times \frac{C_l}{G}}$.
Finally, we concatenate all quantized sub-tokens for reconstruction,
\begin{equation}
\begin{split}
    \mathbf{z}_q(x) = \text{concat}(\mathbf{e}_{1}^{k_{1}}, ..., \mathbf{e}_{G}^{k_{G}}), \\ 
    \quad \text{where} \quad k_i = \arg\min_{\mathbf{e}_{i}^{k} \in \mathcal{E}_i} \|\mathbf{z}(x_{i}) - \mathbf{e}_{i}^k\|_2,
\end{split}
\label{eq: final_output}
\end{equation}
where $i$ is the index of sub-codebook, $k_i$ is the lookup index from the codebook. 

In our final models, for example, the latent dimension $C_l$ is set to 32, the group size $G$ is set to $4$, and the total codebook size is set to $32768$ at most. Thus, the codebook dimension is very small, i.e. $8$, and sub-codebook size ($8192$) is much smaller than existing methods, which alleviates codebook collapse, and representation capacity of $8192^4$ is much larger. With this divide-and-conquer approach, we can achieve the following abilities:
\begin{itemize}
    \item Easier optimization for codebooks. With reduced dimension of each sub-codebook, we can avoid the issue of dead points~\cite{zheng2023cvq}, as shown in ~\cref{fig:deadpoints}.
    \item Larger representation capacity. Traditional VQ-VAE methods~\cite{esser2021vqgan, sun2024llamagen, wang2024emu3} sample a token from the codebook, limiting the representation capacity of the codebook to $K$. In contrast, our multi-group design can easily boost the capacity to $K^{G}$ by over a \textit{billion-fold}, experimentally (see ~\cref{tab:recon_in256}).
    \item Reduced information loss. With a larger dimension, we avoid information loss and improve reconstruction quality.
\end{itemize}

\paragraph{Nested Masking.} 
To avoid sub-codebooks converging to the same one, we propose to mask some sub-tokens in a nested manner~\cite{rippel2014learning, cai2024matryoshka, kusupati2022matryoshka} during training. Specifically, as illustrated in ~\cref{fig:nested}, given a quantized sub-tokens, we randomly sample the number of units to keep $M_{keep} \in {1, ..., G}$ for reconstruction and remove $G-M_{keep}$ last tokens by masking them out. With such a training method, we enforce the encoder to compress the image into the latent in an ordered manner, and the decoder learns to reconstruct the image in a coarse-to-fine manner. 
~\cref{fig: qual_ab_c2f} shows the reconstructed image with nested $M_{keep}$ quantized tokens. The image is reconstructed with finer details when incorporating more quantized sub-tokens. 

\noindent\textbf{Training loss.} Following LlamaGen~\cite{sun2024llamagen}, we utilize a straight-through gradient estimator~\cite{bengio2013estimating}, and employ the commit loss~\cite{van2017neural}, vq-loss~\cite{van2017vqvae}, perceptual loss from LPIPS~\cite{zhang2018unreasonable}, and GAN loss~\cite{isola2017image}, and employ the same weights as~\cite{sun2024llamagen}. Additionally, we employ Charbonnier loss~\cite{lai2018fast, hu2024drivingworld} instead of L1 loss, defined as $L_{Char}=\sqrt{(f(x)-x)^2+\epsilon^2} \approx |f(x)-x|=L_1,$ and L2 loss for reconstruction.The total loss is written as follows:
\begin{equation}
\begin{aligned}
    \mathcal{L}_{\text{total}} &= \lambda_1 \mathcal{L}_{l2} + \lambda_2 \mathcal{L}_{\text{Char}} + \lambda_3 \mathcal{L}_{\text{Commit}} + \lambda_4\mathcal{L}_{\text{vq}} \\
    &+ \lambda_5 \mathcal{L}_{\text{GAN}} + \lambda_6 \mathcal{L}_{\text{Percept}},
\end{aligned}
\end{equation}
where $\lambda_1, \lambda_2, \lambda_3, \lambda_4, \lambda_5$ denote the weights for different losses. Specifically, we set $\lambda_1 = 2, \lambda_2=1, \lambda_3=0.25, \lambda_4=1, \lambda_5=0.5, \lambda_6=1$.

%% file: sec/4_exp.tex
\section{Implementation Details}
\label{sec:exp}

\label{sec:exp_setup}
\paragraph{Training setup.}
The sizes of our image tokenizer models are 308M, 210M, and 122M, corresponding to compression ratios of $32$, $16$, and $8$, respectively. All models are trained with a constant learning rate of $10^{-5}$, a weight decay of $5\times10^{-2}$, and the AdamW~\cite{loshchilov2017decoupled} optimizer ($\beta_1=0.9, \beta_2=0.95$). 
When training the model with group 4, the nested masking probability is $\{0.1,0.1,0.1,0.7\}$.
On the ImageNet benchmark, we strictly follow LlamaGen~\cite{sun2024llamagen} training setting.
On zero-shot benchmarks, we enclose more training data, including FFHQ~\cite{karras2019style}, OpenImages~\cite{kuznetsova2020open}, Segment Anything~\cite{kirillov2023segment}, Argoverse2~\cite{wilson2023argoverse}, Mapillary Vistas~\cite{neuhold2017mapillary}, LAION-HQ~\cite{schuhmann2022laion}.

\paragraph{Evaluation datasets and metrics.}
We evaluate the methods on the ImageNet 50k validation set at a resolution of $256\times256$, and on our collected zero-shot benchmark at resolutions of $512\times512$ and $2560\times1440$. 
Specifically, we construct UHDBench for zero-shot evaluation, which includes $2293$ images at 2k resolution sourced from the ground-truth test sets of HRSOD~\cite{zeng2019towards}, LIU4k~\cite{Liu4K}, UAVid~\cite{lyu2020uavid}, UHDM~\cite{yu2022towards}, and UHRSD~\cite{zeng2019towards}. Images smaller than $2560\times1440$ are excluded.
Furthermore, we collect CelebA~\cite{liu2015celeba}, Davis~\cite{perazzi2016davis}, TextOCR~\cite{singh2021textocr}, VFHQ~\cite{xie2022vfhq}, Spring~\cite{mehl2023spring}, ENeRF~\cite{lin2022enerf} for more zero-shot evaluation. 
The image reconstruction quality is measured by PSNR and reconstruction-Fréchet Inception Distance (rFID). We also report SSIM and LPIPS~\cite{zhang2018unreasonable} to assess reconstruction quality on the ImageNet benchmark.

\section{Experiments}
\subsection{Main Results}
\paragraph{ImageNet Benchmark Reconstruction Comparison.}
We compare \NickName~for two downsample ratios (i.e. $8$ and $16$), with both discrete tokenizers (VQGAN~\cite{esser2021vqgan}, LlamaGen~\cite{sun2024llamagen}, Open-MAGVIT2~\cite{luo2024oplfq}, VAR~\cite{tian2025visual}) and continuous tokenizers (SD-VAE~\cite{rombach2022sd}, SDXL-VAE~\cite{podell2023sdxl} on ImageNet~\cite{deng2009imagenet}. 
In~\cref{tab:recon_in256}, the representation capacity represents the possibility of a token by sampling from the codebook. 
Benefiting from our sub-codebooks design, our \NickName~could achieve larger representation capacity while eliminating the latent dimension downsampling. We achieve the state-of-the-art performance compared to all existing quantization methods.
As reported in~\cref{tab:recon_in256}, \NickName~significantly outperforms the existing discrete tokenizers and even surpasses the continuous tokenizer, SD-VAE~\cite{rombach2022sd} and SDXL-VAE, which have been trained on billions of images.
Furthermore, we compare 2 different groups, G4 ($4$ groups) and G8 ($8$ groups), and their latent dimensions are the same. As G8 has a larger representation capacity, it achieves much better performance. 
\begin{table}[t]
\centering
\setlength{\tabcolsep}{3.5pt} 
\begin{tabular}{lcccc}
\toprule
\multirow{2}{*}{\textbf{Method}} & \multicolumn{2}{c}{\textbf{UHDBench}} & \multicolumn{2}{c}{\textbf{DAVIS}} \\
\cmidrule(lr){2-3} \cmidrule(lr){4-5} &\textbf{rFID}$\downarrow$ & \textbf{PSNR}$\uparrow$  & \textbf{rFID}$\downarrow$ & \textbf{PSNR}$\uparrow$ \\
\midrule
\rowcolor[gray]{.9}SD-VAE~\cite{rombach2022sd}    & \textbf{1.07} & \underline{26.86}& \textbf{1.92} & \underline{28.02}  \\
  
 VQGAN~\cite{esser2021vqgan}        &  5.95 & 22.91& 12.74 & 24.02 \\
 LlamaGen~\cite{sun2024llamagen} & 5.59 & 23.90 & 11.63 & 24.90  \\ 
   OpenMagvit2~\cite{luo2024oplfq}&4.18&23.91 & 7.64 & 25.24   \\
    VAR~\cite{tian2025visual} & 9.85 & 21.79 & 21.85 & 22.52  \\ 
   \NickName-G4& \underline{1.59}&\textbf{28.27}& \underline{2.77}& \textbf{29.50}\\

\bottomrule
\end{tabular}
\caption{\textbf{Zero-shot reconstruction evaluation with a downsample ratio of $16$ on 2560$\times$1440 datasets.} \colorbox[gray]{.9}{Gray} denotes continuous tokenizers. \NickName-G4 indicates our method with $4$ sub-codebooks. }
\label{tab:recon_zero2k}
\vspace{-5mm}
\end{table}

For VQ-based methods, although LlamaGen~\cite{sun2024llamagen} increases the codebook size from $16384$ to $32768$, the performance decreases. 
We conjunct it with the codebook usage decrease. In contrast, ours has the same codebook size ($32768$) but a higher representation capacity and $100\%$ usage, resulting in better performance ($2.26$ v.s. $0.64$).

Additionally, Open-MAGVIT2~\cite{yu2023lfq} adopts a different quantization type, i.e. look-up free quantization (LFQ). The codebook size is much higher, but its representation capacity is much lower than ours. We conjunct that this is why our performance is much higher.

\begin{figure*}[t]
\begin{center}
\vspace{-5pt}
\centerline{\includegraphics[width=2.0\columnwidth]{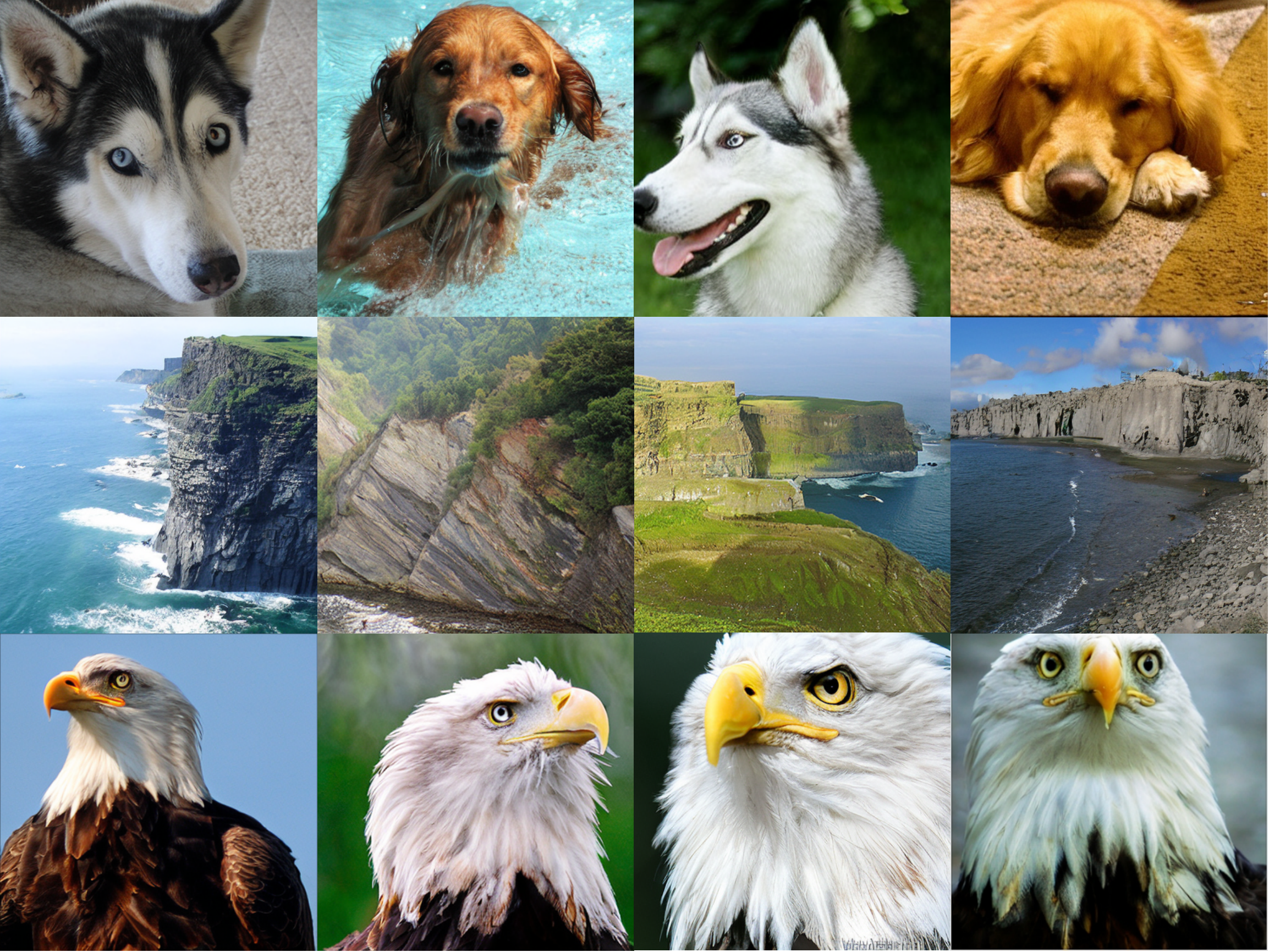}}
\caption{\textbf{Visualization of class-to-image generation.} The classes are dog(Golden Retriever and Husky), cliff, and bald eagle of ImageNet-1k dataset, for each row.}
\label{fig:c2i_gen}
\end{center}
\vspace{-15pt}
\end{figure*}

\noindent\textbf{Zero-shot Generalization Reconstruction Comparison.}
To validate the generalization ability, we construct $8$ zero-shot benchmarks.
We conduct experiments with both $2560\times 1440$ resolution in~\cref{tab:recon_zero2k} on $2$ benchmarks and  $512\times 512$ resolution in~\cref{tab:recon_zero512} on $6$ benchmarks.
Our performance ranks the first on both rFID and PSNR metrics. Note that we even outperform SD-VAE~\cite{rombach2022sd} significantly on PSNR. Furthermore, we prepare UHDBench and DAVIS for 2K images reconstruction evaluation. Results are reported in~\cref{tab:recon_zero2k}. Our method still achieves the best performance on PSNR in 2K image reconstruction. Thanks to our sub-codebooks design, our \NickName~effectively captures the enriched features. 

\noindent\textbf{Image Generation Results.}
We strictly follow LlamaGen's code~\cite{sun2024llamagen} and settings in the ImageNet generation experiment, but replace its VQ tokenizer with ours. We apply $cfg=2.0$ and $topk=4096$ for each sub-codebook for image generation. As shown in~\cref{tab:t_gen}, our generation quality is better than the baseline, especially on IS. Visualization of samples are illustrated in~\cref{fig:c2i_gen}, showing that \NickName is able to generate diverse class-conditional images with high fidelity, reality, and quality. Detailed textures are accurately recovered with ours VQ tokenizer.
\begin{table}[t]
    \centering
    \resizebox{.8\linewidth}{!}{
    \begin{tabular}{lcccc} 
    \toprule
    \multicolumn{1}{l}{Type}    &
    \multicolumn{1}{c}{Methods}               
     & \multicolumn{1}{c}{Params} & \multicolumn{1}{c}{gFID$\downarrow$} & \multicolumn{1}{c}{IS$\uparrow$}  \\
     \midrule
    \midrule
    VQ & VQGAN~\cite{esser2021vqgan}  & 1.4B & 15.78 & 74.3 \\
    VQ & RQ-VAE~\cite{lee2022autoregressive} & 1.4B &  8.71 & 119.0 \\
    VQ & LlamaGen~\cite{sun2024llamagen} & 1.4B &  3.09 & 253.6\\
    VQ & ours & 1.4B &  \textbf{3.02} & \textbf{294.1} \\
    \bottomrule
    \end{tabular}
    }

    \caption{\textbf{ImageNet 256p class-to-image generation evaluation.} }
    \label{tab:t_gen}
\end{table}

\noindent\textbf{Qualitative Comparison of Image Tokenizers.}
As shown in~\cref{fig:qual}, we present a qualitative comparison of \NickName~with SD-VAE~\cite{rombach2022sd}, LlamaGen~\cite{sun2024llamagen}, and OpenMagvit2~\cite{luo2024oplfq} on our collected UHDBench. Compared to others, \NickName~displays a significant improvement in reconstructing high fidelity details, especially for complex textures and tiny texts, surpassing both discrete and continuous tokenizers.
In particular, LlamaGen~\cite{sun2024llamagen} fails to reconstruct fine details. In row $3$, the letters' color is falsely reconstructed from green to white. We conjecture that this is caused by the limited latent representation capacity. 
By extending the codebook size, OpenMagvit2~\cite{luo2024oplfq} enhances the ability for more realistic colors and details than LlamaGen~\cite{sun2024llamagen}. However, there remains a failure in accurately reconstructing tiny letters.
Note that SD-VAE~\cite{rombach2022sd}, with a continuous latent space, has the highest ability for reconstruction among other VQ-VAEs. It achieves a relatively good reconstruction of details. However, the edges are blurred and the reconstruction of tiny details is ineffective.
In contrast, ours \NickName~preserves latent dimension to retain more encoded features and learns a coarse-to-fine representation by sub-codebooks, enabling us to reconstruct these tiny textures accurately.

\begin{table}[t]
\centering
\resizebox{.95\linewidth}{!}{
\begin{tabular}{ccccccc}
\toprule
\textbf{SC} & \textbf{NM} & \textbf{rFID}$\downarrow$ & \textbf{PSNR}$\uparrow$ & \textbf{SSIM}$\uparrow$ & \textbf{LPIPS}$\downarrow$ & \textbf{Usage}$\uparrow$ \\
\midrule
& & 2.49& 20.49& 0.599 & 0.212 & 80\% \\
$\checkmark$ & & \underline{0.75}&  \underline{23.41} & \underline{0.747}& \underline{0.119} & 93\%\\
$\checkmark$ & $\checkmark$&  \textbf{0.64} & \textbf{23.71}& \textbf{0.755}& \textbf{0.110} & 100\% \\
\bottomrule
\end{tabular}
}
\caption{\textbf{Ablation study of proposed strategies.} SC indicates sub-codebooks. NM indicates nested masking. }
\vspace{-15pt}
\label{tab:ab_subgroup}

\end{table}
\begin{table}[t]
\centering
\setlength{\tabcolsep}{4.0pt} 

\begin{tabular}{lcccc}
\toprule
\textbf{Method} & \textbf{rFID}$\downarrow$ & \textbf{PSNR}$\uparrow$ & \textbf{SSIM}$\uparrow$ & \textbf{LPIPS}$\downarrow$ \\
\midrule
Ours ($M_{keep}=1$)  & 196.43& 12.83& 0.194 & 0.602 \\
Ours ($M_{keep}=2$)  & 80.00& 15.55& 0.369 & 0.438 \\
Ours ($M_{keep}=3$) &  \underline{4.56} & \underline{20.99}& \underline{0.629}& \underline{0.202} \\
Ours ($M_{keep}=4$)&  \textbf{0.64} & \textbf{23.71}& \textbf{0.755}& \textbf{0.110} \\
\bottomrule
\end{tabular}
\caption{\textbf{Ablation study of nested masking. }$M_{keep}$ indicates the number of used groups to keep in the nested masking strategy.}
\label{tab:ab_c2f}
\vspace{-15pt}
\end{table}

\subsection{Ablation Study}
\subsubsection{Ablation on nested masking and sub-codebooks}
This ablation study investigates the contribution of two key factors: the nested masking strategy and the sub-codebook design.
A codebook size of $32768$ is maintained throughout, and for models using sub-codebooks, the group parameter $G=4$.
We ablation on ImageNet~\cite{deng2009imagenet}.

As shown in~\cref{tab:ab_subgroup}, the model without sub-codebooks shows lower PSNR and higher rFID, probably due to its limited representation capacity of only $2^{15}$.
When the nested masking is applied, a clear improvement in reconstruction fidelity is observed, confirming the strategy’s effectiveness.

~\cref{tab:ab_c2f}~investigates how the nested masking works. Specifically, 
by analyzing different values of $M_{keep}$, each sub-codebook’s impact can be isolated. 
We can conclude that consistent performance gains when additional sub-codebooks are utilized.
The nested masking process enables a \textit{coarse-to-fine reconstruction}. 
As shown in ~\cref{fig: qual_ab_c2f}, even with one fewer sub-codebook, coarse textures can still be recovered, while using all sub-codebooks enables finer details. This suggests that the encoder is guided to organize latent representations in a hierarchical manner, and the decoder leverages this hierarchy to progressively refine the output.

\begin{figure}[t]
\begin{center}
\vspace{-20pt}
\centerline{\includegraphics[width=0.99\columnwidth]{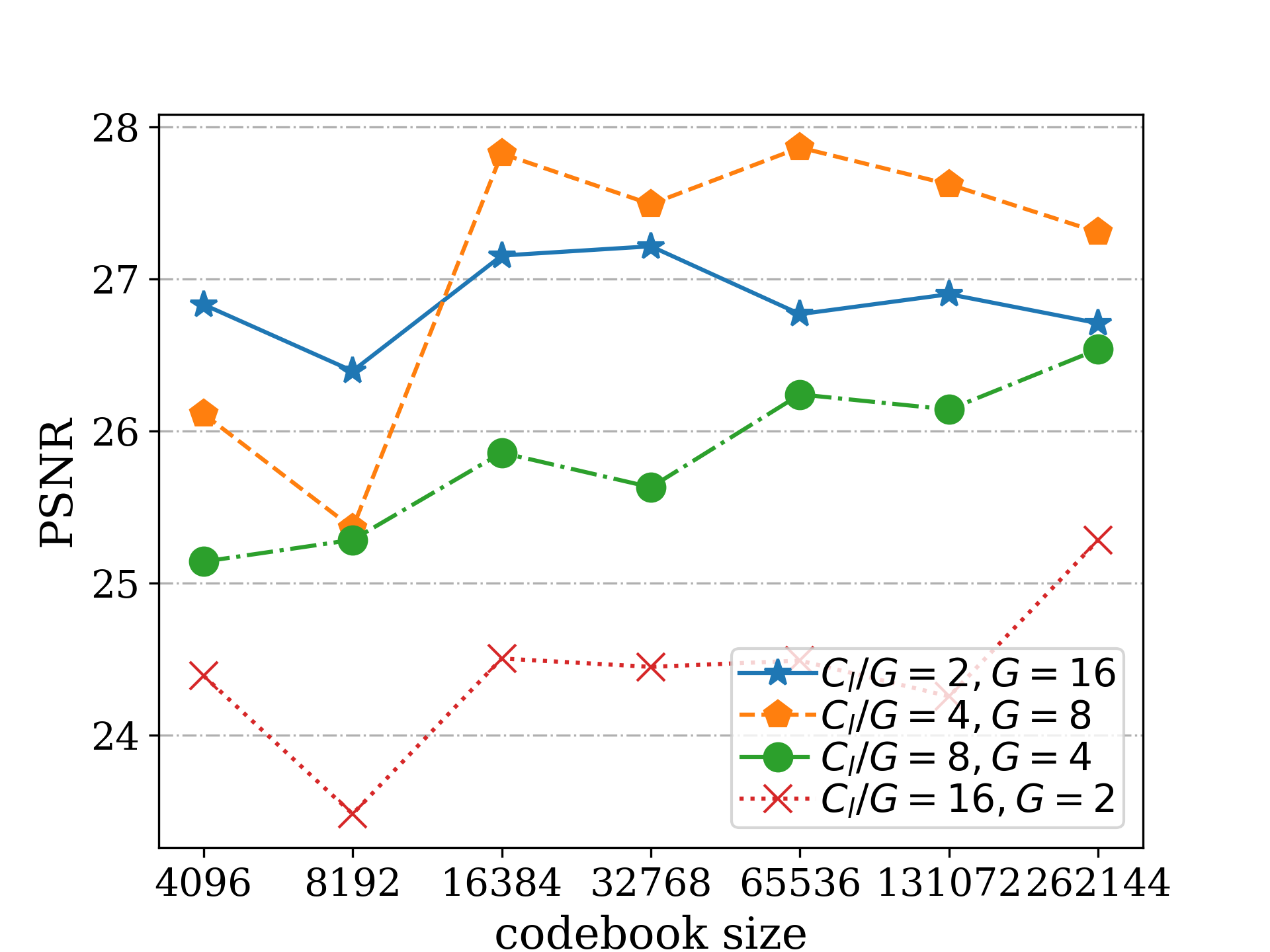}}
\caption{\textbf{Reconstruction quality of different latent dimensions and codebook size.} $G$ indicates the number of sub-codebooks. $C_l$ indicates the latent dimension of the encoded features and is fixed at $32$. $C_l/G$ indicates the dimension of each sub-token.}
\label{fig:ab_dim_size}
\end{center}
\vspace{-35pt}
\end{figure}

\noindent\textbf{Ablation on sub-codebook number, sub-token dimension, and codebook size}
We ablate on ImageNet. To decouple the encoding, we fix the encoded latent dimension $C_l=32$. The number of sub-codebooks $G$ varies from $2$ to $16$, and the sub-token dimension $C_l/G$ is adjusted from $16$ to $2$. The codebook size varies from $4096$ to $262144$.~\cref{fig:ab_dim_size} shows the PSNR of different settings. 
With the same codebook size, the performance is better with increasing sub-codebook groups number and decreasing sub-codebook dimensions.
This is because larger sub-codebooks group numbers improve the representation capacity, while lower sub-codebook dimensions ease the optimization.
However, when the sub-codebook groups number increases to $16$, the performance is worse than that of $8$. We conjunct that this is because each sub-token dimension shrinks to 2, reducing each code's representation ability. 
As shown in~\cref{fig:ab_dim_size}, when the codebook size is lower than $16384$, decreasing the codebook size will cause inferior performance. However, when the codebook size is larger than 16384, the performance will saturate.

%% file: sec/5_conclusion.tex
\section{Conclusion}
\label{sec:conclusion}

In conclusion, \NickName~enhances VQ-VAE performance by expanding the representation capacity of discrete codebooks with reduced information loss. 
By preserving the latent dimension and employing multiple sub-codebooks, \NickName~addresses the limitations and narrows the gap between VQ-VAEs and VAEs. 
Experimental evidence on both ImageNet 256p and zero-shot HD benchmarks confirms its state-of-the-art reconstruction quality, surpassing strong baselines such as SD-VAE in key metrics. 
These contributions highlight the capability of \NickName~in HD image compression and reconstruction and facilitate future research into more efficient discrete latent representations and downstream tasks.